\def\year{2022}\relax
\documentclass[letterpaper]{article} 
\usepackage{aaai22}  
\usepackage{times}  
\usepackage{helvet}  
\usepackage{courier}  
\usepackage[hyphens]{url}  
\usepackage{graphicx} 
\usepackage{natbib}  
\usepackage{caption} 
\DeclareCaptionStyle{ruled}{labelfont=normalfont,labelsep=colon,strut=off} 
\usepackage{algorithm}
\usepackage{algorithmic}
\usepackage{pdfpages}
\usepackage{newfloat}
\usepackage{listings}
\floatstyle{ruled}
\newfloat{listing}{tb}{lst}{}
\floatname{listing}{Listing}

\usepackage{amsmath}
\usepackage{amsfonts}
\usepackage{pifont}
\title{Resistance Training using Prior Bias: toward Unbiased Scene Graph Generation}
\author {
	Chao Chen \textsuperscript{\rm 1,2}\footnote{This work was done when Chao Chen was a research intern at JD Explore Academy.},
	Yibing Zhan \textsuperscript{\rm 2}, 
	Baosheng Yu \textsuperscript{\rm 3}, 
	Liu Liu \textsuperscript{\rm 3}, 
	Yong Luo \textsuperscript{\rm 1}\footnote{Corrsponding Author.}, 
	Bo Du \textsuperscript{\rm 1}\footnotemark[2]
}
\affiliations {
	\textsuperscript{\rm 1}  National Engineering Research Center for Multimedia Software, Institute of Artificial Intelligence, School of Computer Science and Hubei Key Laboratory of Multimedia and Network Communication Engineering, Wuhan University, China, \\
	\textsuperscript{\rm 2} JD Explore Academy, China, 
	\textsuperscript{\rm 3} The University of Sydney, Australia, \\
	\{chenchao, luoyong, dubo\}@whu.edu.cn,\\ zhanyibing@jd.com, baosheng.yu@sydney.edu.au, liu.liu1@sydney.edu.au,
}

\begin{document}

\maketitle

\begin{abstract}
	\textbf{S}cene \textbf{G}raph \textbf{G}eneration (SGG) aims to build a structured representation of a scene using objects and pairwise relationships, which benefits downstream tasks. However, current SGG methods usually suffer from sub-optimal scene graph generation because of the long-tailed distribution of training data. To address this problem, we propose Resistance Training using Prior Bias (RTPB) for the scene graph generation. Specifically, RTPB uses a distributed-based prior bias to improve models' detecting ability on less frequent relationships during training, thus improving the model generalizability on tail categories. In addition, to further explore the contextual information of objects and relationships, we design a contextual encoding backbone network, termed as Dual Transformer (DTrans). We perform extensive experiments on a very popular benchmark, VG150, to demonstrate the effectiveness of our method for the unbiased scene graph generation. In specific, our RTPB achieves an improvement of over 10\% under the mean recall when applied to current SGG methods. Furthermore, DTrans with RTPB outperforms nearly all state-of-the-art methods with a large margin.
	\footnote{Code will be available at https://github.com/ChCh1999/RTPB}

\end{abstract}
\section{Introduction}

\begin{figure}[!ht]
	\centering
	\includegraphics[width=\linewidth]{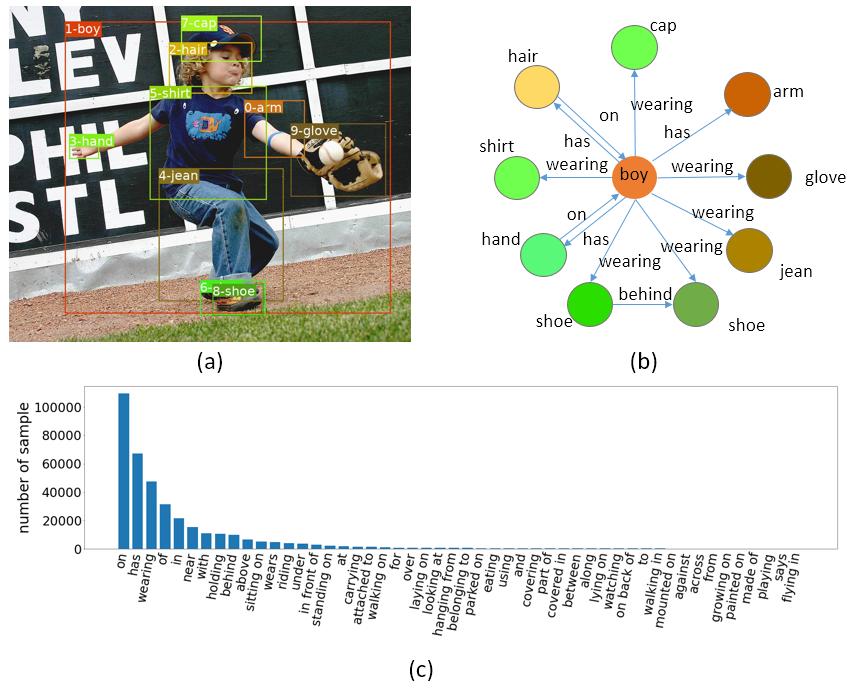} %
	\caption{(a) an input image with detected objects. (b) a generated scene graph, which is a graphical representation of the input image with objects and pairwise relationships. (c) the long-tail distribution of object relationships in the Visual Genome (VG) dataset~\cite{krishna2016visual}.}
	\label{fig:sgg_sample}
\end{figure}

Scene graph generation aims to understand the semantic content of an image via a scene graph, where nodes indicate visual objects and edges indicate pairwise object relationships. An intuitive example of scene graph generation is shown in Figure~\ref{fig:sgg_sample}. Scene graph generation is beneficial to bridge the gap between the low-level visual perceiving data and the high-level semantic description. Therefore, a reliable scene graph can provide powerful support for downstream tasks, such as image captioning~\cite{Zhong2020Comprehensive}, image retrieval~\cite{scene-graph}, and visual question answering~\cite{VCTree_tang}. 

Scene graph generation usually suffers from the long-tail problem of relationships in training data \cite{Tang2020_unbiased,chen2019knowledge,zhan2019exploring}. 
For example, as shown in Figure~\ref{fig:sgg_sample}(c), the widely used scene graph generation dataset, Visual Genome ~\cite{krishna2016visual}, is dominated by a few relationships with coarse-grained descriptions (head categories), whereas there are no sufficient annotations available for other less frequent relationships (tail categories). This severe imbalanced distribution makes it difficult for an unbiased scene graph generation, considering there are only a few training samples in real-world scenarios for recognizing less frequent relationships.

To address the above-mentioned issues, many class re-balancing strategies have been introduced for unbiased scene graph generation~\cite{Tang2020_unbiased,li2021bipartite}. However, existing methods still struggle to achieve satisfactory performance on tail categories, and more sophisticated solutions are desirable. 
Inspired that humans increase muscle strength by making their muscles work against a force, we propose resistance training using prior bias (RTPB) to improve the detection of less-frequently relationships and address the long-tail problem in SGG.
Specifically, RTPB assigns a class-specific resistance to the model during training, where the resistance on each relationship is determined by a prior bias, named resistance bias. The resistance bias is first initialized by the prior statistic about the relationships in the training set. The motivation of a resistance bias is to enforce the model to strengthen the ability on less frequent relationships, since they are always corresponding to heavier resistance. In this way, RTPB enables the model to resist the influence of the imbalanced training dataset.
Furthermore, to better explore global features for recognizing the relationships, we design a contextual encoding backbone network, named dual Transformer (DTrans for short). Specifically, the DTrans uses two stacks of the Transformer to encode the global information for objects and relationships sequentially. 
To evaluate the effectiveness of the proposed RTPB, we introduce it into recent state-of-the-art methods and our DTrans. Specifically, RTPB can bring an improvement of over 10\% in terms of the mean recall criterion compared with other methods. By integrating with the DTrans baseline, RTPB achieves a new state-of-the-art performance for the unbiased scene graph generation.

The main contributions of this paper are summarized as follows:
1) We propose a novel resistance training strategy using prior bias (RTPB) to improve the model generalizability on  the rare relationships in the training dataset;
2) We define a general form of resistance bias and devise different types of specific bias by using different ways to model object relationship;
3)   We introduce a contextual encoding structure based on the transformer to form a simple yet effective baseline for scene graph generation.
Extensive experiments on a very popular benchmark demonstrate significant improvements of our RTPB and DTrans compared with many competitive counterparts for unbiased SGG.

\section{Related Work}

\subsection{Scene Graph Generation}
Scene graph generation belongs to visual relationship detection and uses a graph to represent visual objects and the relationships between them~\cite{scene-graph}. In early works, detection and prediction are performed only based on specific objects or object pairs~\cite{Lu16}, which can't make full use of the semantic information in the input image. To take the contextual information into consideration, later works utilize the whole image by designing various network architectures, such as biLSTM~\cite{Zellers_2018_CVPR}, TreeLSTM~\citep{VCTree_tang} and GNN~\cite{Yang_2018_ECCV,li2021relationship}.
Some other works try to utilize external information, such as linguistic knowledge and knowledge graph, to further improve scene graph generation~\cite{Lu16, chen2019knowledge, Yu2017,gu2019scene}.

However, due to the annotator preference and image distribution, this task suffers from a severe data imbalance issue. Dozens of works thus try to address this issue, and 
the main focus is on reducing the influence of long-tail distribution and building a balanced model for relationship prediction. Tang et al. also try to use causal analysis \cite{Tang2020_unbiased} to reduce the influence of training data distribution on the final model. Some other works \cite{chen2019soft,zhan2020multi,chiou2021recovering} address this issue in a positive-unlabeled learning manner, and typical imbalance learning methods, such as re-sampling and cost-sensitive learning, are also introduced for scene graph generation~\cite{li2021bipartite, yan2020pcpl}. Unlike these approaches, we adopt the resistance training strategy using prior bias (RTPB), which utilizes a resistance bias item for the relationship classifier during training to optimize the loss value and the classification margin of each type of relationship.

\subsection{Imbalanced Learning}

Imbalanced learning methods can be roughly divided into two categories: resampling and cost-sensitive learning.

\subsubsection{Resampling} Resampling methods change the class ratio to make the dataset a balanced one. These methods can be divided into two categories: oversampling and undersampling. Oversampling increases the count of the less frequent classes~\cite{chawla2002smote},
and hence may lead to overfitting for minority classes. Undersampling reduces the sample of major categories, and thus is not feasible when data imbalance is severe since it discards a portion of valuable data~\cite{liu2008exploratory}.


\subsubsection{Cost-sensitive Learning}
Cost-sensitive learning assigns different weights to samples of different categories \cite{elkan2001foundations}.
Re-weighting is a widely used cost-sensitive strategy by using prior knowledge to balance the weights across categories. Early works use the reciprocal of their frequency or a smoothed version of the inverse square root of class frequency to adjust the training loss of different classes.
However, this tends to increase the difficulty of model optimization under extreme data imbalanced settings and large-scale scenarios. \citeauthor{cui2019class}
propose class balanced (CB) weight, which is defined as the inverse effective number of samples\cite{cui2019class}. Some other approaches dynamically adjust the weight for each sample based on the model's performance~\cite{Lin_2017_ICCV, Li_Liu_Wang_2019, cao2019learning}.

Different from these approaches, which directly deal with the loss function, we add a prior bias on the classification logits for each class based on the distribution of training dataset. In this way, we provide a new approach to import additional category-specific information and address the imbalance problem through adjusting the classification boundaries.
A similar idea is proposed in~\cite{menon2020long} for long-tailed recognition, and we differ from them significantly in the tasks and formulations.
\section{Methods}
\begin{figure*}[ht!]
	\centering
	\includegraphics[width=0.79\linewidth]{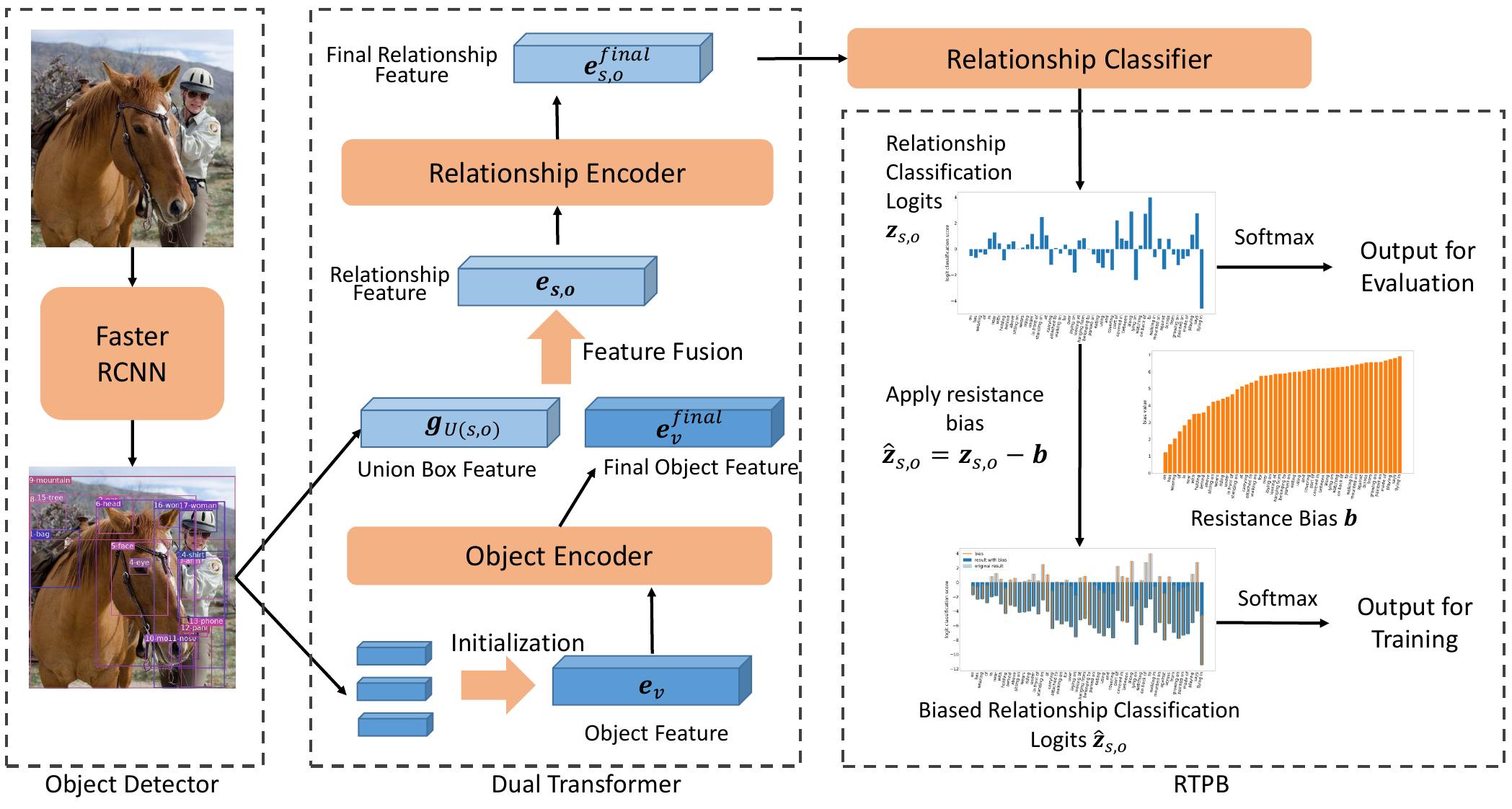} 
	\caption{The overall structure of our method. We adopt Faster RCNN as the object detector. Dual Transformer is used to encode object/relationship features. RTPB addresses the long-tail problem by adjusting the classification logits of each relationship.}
	\label{fig:model}
\end{figure*}
\subsection{Problem Setting and Overview}
\subsubsection{Problem Setting}
Given an image $\mathcal{I}$, our goal is to predicate a graph $\mathcal{G} = \{\mathcal{V},\mathcal{E}\}$, where $\mathcal{V}$ is the set of objects in the image and $\mathcal{E}$ is pairwise relationships of objects in $\mathcal{V}$. Each object $v \in \mathcal{V}$ consists of the bounding box coordinates $\textbf{\textit{b}}_v$ for location and the class label $c_v$ for type. An edge $e \in \mathcal{E}$ should include the pair of a subject and a object, \textit{i.e.}, $(v_s, v_o)$,  and the label of the relationship $r$ between $v_s$ and $v_0$.
Typically, for object $v\in \mathcal{V}$, the label  $c_v$ belongs to $\mathcal{C}_e = \{1,2,...,L_e\}$, where $L_e$ is the number of object labels. The label of relationship $c_r$ are in $\mathcal{C}_r = \{1,2,...,L_r\}$, where $L_r$ is the number of relationship labels. 
\par In the commonly used SGG pipeline, the probability of scene graph $Pr(\mathcal{G}|\mathcal{I})$ is formulated as follows
\begin{equation}
  Pr(\mathcal{G}|\mathcal{I}) = Pr(\mathcal{B}|\mathcal{I})Pr(\mathcal{O}|\mathcal{I},\mathcal{B})Pr(\mathcal{G}|\mathcal{I},\mathcal{B}, \mathcal{O}),  
\end{equation}
where $ Pr(\mathcal{B}|\mathcal{I})$ indicates the proposal generation. The $\mathcal{B}$ is the set of bounding box $b_v$, which is conducted by an object detector. $Pr(\mathcal{O}|\mathcal{I},\mathcal{B})$ denotes the object classification. The $\mathcal{O}$ is the set of object class $c_v$, and $Pr(\mathcal{G}|\mathcal{I}, \mathcal{B}, \mathcal{O})$ means the final relationship classification.

\subsubsection{Methods Overview}
Figure \ref{fig:model} shows the overall structure of our method. Following previous works \cite{Tang2020_unbiased,Zellers_2018_CVPR}, we utilize Faster RCNN \cite{ren2015faster} to generate object proposals and the corresponding features from the input image.
For each object proposal, the object detector conducts the bounding box coordinates 
$\textbf{\textit{b}}_v$, 
the visual feature 
$\textbf{\textit{g}}_v$,
and the object classification score $\textbf{\textit{z}}_v\in\mathbb{R}^{|\mathcal{C}_v|}$.
Besides, to improve the relationship prediction, the backbone also generates the feature of the union boxes of each pair of objects.
Then, we introduce our \textit{Dual Transformer (DTrans)}, which uses two stacks of Transformers to encode the context-aware representations for objects and relationships, respectively (see Section \ref{sec:DTrans}). Moreover, to address the long-tail problem for SGG, we adopt the \textit{\textbf{r}esistance \textbf{t}raining using \textbf{p}rior \textbf{b}ias (RTPB)} for unbiased SGG. The RTPB assigns resistance to the model by the prior bias, named resistance bias (see Section \ref{sec:RTPB}).
%


\subsection{Dual Transformer}
\label{sec:DTrans}
We propose the Dual Transformer (DTrans) for better contextual information encoding. As shown in Figure \ref{fig:model}, based on the outputs of the object detector, i.e., the location of objects, the feature of objects, and the feature of union boxes, the DTrans uses self-attention to encode the context-aware representation of objects and relationships.
%

\subsubsection{Self-attention}
Our model uses two stacks of transformers to obtain the contextual information for objects and the corresponding pairwise relationships. Each transformer encodes the input features with self-attention mechanisms \cite{vaswani2017attention}, which mainly consists of the attention and the feed-forward network. The attention matrix is calculated as
\begin{equation}
	Attention(Q,K,V) = \sigma(\frac{QK^T}{\sqrt{d_k}})V,
\end{equation}
where query ($Q$), keys ($K$), and value ($V$) are obtained from the input feature through three different linear transformation layers, $1/\sqrt{d_k}$ is the scaling factor for the dot product of $Q$ and $K$~\cite{vaswani2017attention}, and $\sigma$ is the softmax function. Besides, we utilize the multi-head attention, which is used to divide the Q, K, and V into $n_h$ parts and calculates the attentions of each part. Next, a feed-forward network (FFN) is used to mix the attentions of each part up.

\subsubsection{Object Encoder}
As shown in Figure \ref{fig:model}, based on all the results about object proposals from the object detector, we first compose them to initialize a feature vector of the corresponding object as follows:
\begin{equation}
	\mathbf{e}_v = W_o[pos(\textbf{\textit{b}}_v),\textbf{\textit{g}}_v, ebd(c_v)],
\end{equation}
where $[\cdot,\cdot]$ denotes the concatenation operation, the $pos$ is a learnable position encoding method for object location, $ebd(c_v)$ is the GloVe vector embedding for object label $c_v$, and $W_o$ is a linear transformation layer used to initialize the object feature from the above information. 
Then, we feed the $\mathbf{e}_v$ into a stack of $n_o$ transformers to obtain the final context-aware features $\mathbf{e}_v^{final}$ for objects. 
The $\mathbf{e}_v^{final}$ is first used to produce the final classification result of objects as follows:
\begin{equation}
    \mathbf{p}_v = \sigma(W_{clf}^{o}\mathbf{e}_v^{final}),
\end{equation}
where $W_{clf}^{o}$ is the object classifier. 
And then, the $\mathbf{e}_v^{final}$ is also used to generate the features of the pairwise relationships of objects.

\subsubsection{Relationship Encoder}
Before encoding the features of relationships, we first carry out a feature fusion operation to generate the basic representation of relationships. For each directed object pair $(s,o)$, we concatenate the visual feature $\mathbf{g}_{U(s,o)}$ of the union box of $s$ and $o$, and the context-aware features of the subject $s$ and the object $o$. Then, we use a linear transformation to obtain the representation of the relationship between the subject $s$ and the object $o$:
\begin{equation}
	\mathbf{e}_{s,o} = W_r[\mathbf{g}_{U(s,o)},\mathbf{e}_s^{final}, \mathbf{e}_o^{final}],
\end{equation}
where $W_r$ is a linear transformation layer used to compress the relationship features. $\mathbf{e}_s^{final}$ and $ \mathbf{e}_o^{final}$ are the final context-aware features of subject $s$ and object $o$.

Then, we use another stack of $n_r$ Transformers to encode the features of relationships and produce the final feature $\mathbf{e}_{s,o}^{final}$ for the relationship between subject $s$ and object $o$. Finally, the relationship classifier uses the feature $\mathbf{e}_{s,o}^{final}$ to recognize the pairwise relationships as follows
\begin{equation}
	\mathbf{p}_{s,o} = \sigma(\mathbf{z}_{s,o}) = \sigma(W_{cls}^r\mathbf{e}_{s,o}^{final}),
	\label{eq:p}
\end{equation}
where $\mathbf{z}_{s,o} = W_{cls}\mathbf{e}_{s,o}^{final}$ is  vector of the logits for relationship classification, and$W_{cls}^r$ is the relationship classifier.

\subsection{Resistance Training using Prior Bias}
\label{sec:RTPB}
Inspired by the idea of resistance training that muscle turns to be stronger if they are training with heavier resistance \cite{kraemer2004fundamentals},
we propose the resistance training using prior bias (RTPB) for unbiased SGG. Our RTPB uses the prior bias as the resistance for the model only during training to adjust the model's strength for different relationships. We name the prior bias as resistance bias, which is applied to the model in the training phase as:
\begin{equation}
	\hat{\mathbf{z}}_{s,o} = \mathbf{z}_{s,o} - \mathbf{b},
\end{equation}
where $\mathbf{z}_{s,o}= [z_1,z_2,...,z_{L_r}]$ is the vector of logits for relationship classification in Eq. \eqref{eq:p}, and $\mathbf{b} = [b_1, b_2,...,b_{L_r}]$ is the vector of resistance bias for each relationship. 
By assigning relatively heavier resistance to the tail relationships during training, we enable the model without resistance to handle the tail relationships better and obtain a balanced performance between the tail relationships and the head relationships.

In the following part of this section, 
we first introduce four resistance biases from the prior statistic about the training set. Then, we analyze how RTPB works through the loss and the objective.
\subsubsection{Instances for Resistance Bias}
\par We define the basic form of resistance bias $b_i$ for relationship $i \in \mathcal{C}_r$ as follows:
\begin{equation}
	b_i = -log \left(\frac{w_i^{a}}{\sum_{j \in \mathcal{C}_r} w_j^a}+ \epsilon\right),
	\label{eq:bias}
\end{equation}
where $w_i$ is the weight of relationship $i$, and $a$, $ \epsilon$ are the hyper-parameters used to adjust the distribution of $b_i$. For resistance bias, the category weight $w_i$ is negatively correlated to the  resistance bias value $b_i$. Therefore, to assign the head categories of relationship small resistance, the distribution of the weight $w_i$ should be correlated to the long-tail distribution of the training set. We introduce four resistance biases:

\textit{\textbf{C}ount Resistance \textbf{B}ias (\textbf{CB})} Intuitively, for the general classification task, we can set the bias weight $w_i$ of resistance bias with the proportion of sample for each relationship in the training set. We denote this type of resistance bias as \textbf{c}ount resistance \textbf{b}ias (CB).  

\begin{figure}[t]
	\centering
	\includegraphics[width=0.99\columnwidth]{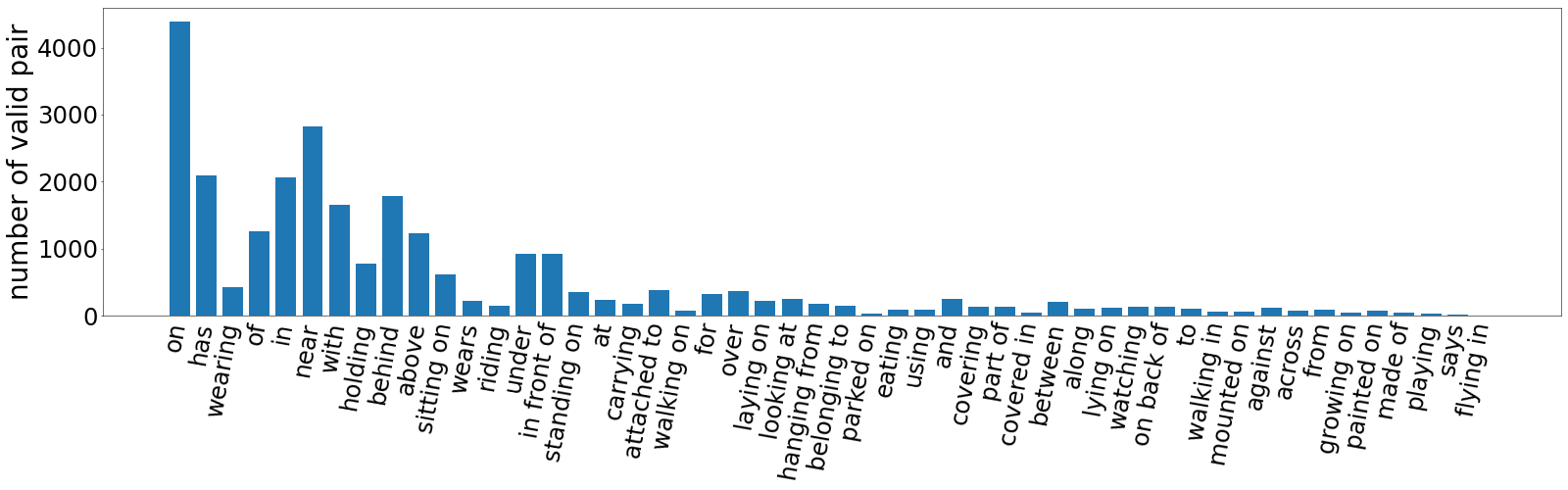} 
	\caption{The distribution of the valid pair count.}
	\label{fig:dist}
\end{figure}

\textit{\textbf{V}alid Resistance \textbf{B}ias (\textbf{VB})} 
Besides the number of samples, the long-tail problem also occurs between the general descriptions and the detailed ones. 
Therefore, we formulate valid pair as follows: 
if there is relationship $i\in \mathcal{C}_r$ for object pair $(s,o)$ in the training set, the $(s,o)$ is a valid pair of relationship $i$. 
The distribution of valid pair count for the popular SGG data set, Visual Genome \cite{krishna2016visual}, is shown in the Figure.\ref{fig:dist}. We set the weight $w_{i}$ as the proportion of valid object pair for relationship $i$. In this way, the RTPB can distinguish between the general descriptions of relationships and the rare ones. We call this resistance bias \textbf{V}alid Resistance \textbf{B}ias (VB for short).

\textit{\textbf{P}air Resistance \textbf{B}ias (\textbf{PB})} However, for the SGG task, the classification of relationship is related to not only the relationship, but the pair of entities. Thus we try to use the relationship distribution of each object pair. For different object pairs, the bias are different. In application, we select the bias based on the classification result of objects. For object pair $(s,o)$, the resistance bias for relationship $i$ is follows:
\begin{equation}
    b_{s,o,i} = -log \left(\frac{w_{s,o,i}^{a}}{\sum_{j \in \mathcal{C}_r} w_{s,o,j}^a}+ \epsilon\right),
    \label{rtpb:eq:bsoi}
\end{equation}
where $w_{s,o,i}$ s the ratio of the relationship $i$ in all the relationships between subject $s$ and object $o$ in the training set. And we call this resistance bias as \textbf{P}air Resistance \textbf{B}ias (PB)

\textit{\textbf{E}stimated  Resistance \textbf{B}ias (\textbf{EB})}  Because many pairs of $s,o$ have only a little number of valid relationships and few annotations, the number of relationship sample for particular object pair can hardly show the general distribution of relationship in many cases. Thus we propose subject-predicate and predicate-object count $n^{sppo}$ to estimate the relationship distribution. For subject $s$, object $o$ and relationship $i$, it's calculated as
\begin{equation}
	n_{s,o,i}^{sppo}  = \sqrt{\sum_{o' \in \mathcal{C}_e}{n_{s,o',i}}\times \sum_{s' \in \mathcal{C}_e}{n_{s',o,i}}},
\end{equation}
where $n_{s,o,i}$ is the count of relationship $i$ between subject $s$ and object $o$. Then we produce a new resistance bias by replacing the weight $w_{s,o,i}$ in Eq.\eqref{rtpb:eq:bsoi} with the proportion ${n_{s,o,i}^{sppo}}/{\sum_{j \in \mathcal{C}_r}{n_{s,o,j}^{sppo}}}$.
We name this type of resistance as \textbf{E}stimated  Resistance \textbf{B}ias (EB)

\subsubsection{Loss with RTPB}
In this section, we analyze the effect of RTPB through classification loss. In the classifier, we conduct the final relationship classification probability by a softmax function, and we use the cross-entropy (CE) to evaluate the classification optimization objective. 
The  classification probability and CE loss are shown in follows,
\begin{align}
 p_i & = Pr(i|s,o,\mathcal{I}) = \sigma (\mathbf{z}_{s,o})_i = \frac{e^{z_{i}}}{\sum_{j\in\mathcal{C}_r}e^{z_{j}}}, \label{eq:p_i}\\
\mathcal{L}_i & = -log(p_i) = -log \frac{e^{z_{i}}}{\sum_{j\in \mathcal{C}_r}e^{z_{j}}}, \label{eq:l_i}
\end{align}
where $p_i$ is the probability that the relationship between object pair $(s,o)$ in image $\mathcal{I}$ is $i$, and $\mathbf{z}_{s,o} $is the vector of logits for relationship classification mentioned in Eq. \eqref{eq:p}. If we apply our RTPB, the probability $p_i$ in Eq. \eqref{eq:p_i} turns to be:
\begin{equation*}
	\hat{p}_i = \sigma ({\mathbf{z}}_{s,o} - \mathbf{b})_i = p_i\frac{e^{ -b_{i}}}{\sum_{j\in \mathcal{C}_r}e^{-b_{j}} p_j},
\end{equation*}
where $\mathbf{b}= [b_1,b_2,...,b_{L_r}]$ is vector of resistance bias.
And the softmax cross-entropy loss turns to be:
\begin{align}
    \mathcal{\hat{L}}_i 
	& = -log(\hat{p}_i)
		  = -log \frac{e^{z_{i} - b_{i}}}{\sum_{j\in \mathcal{C}_r}e^{z_{j} - b_{j}}}
    \nonumber\\
	& = -log \frac{e^{z_i}}{\sum_{j\in \mathcal{C}_r}e^{z_j}} + b_i + log \sum_{j\in\mathcal{C}_r}\frac{e^{-b_j}e^{z_j}}{\sum_{k\in \mathcal{C}_r}e^{z_k}} 
	\nonumber\\
	& = \mathcal{L}_i + \theta_{i}, 
	\label{eq:biasloss}	  
\end{align}
where $~ \theta_{i}  =  b_{i}  + log \sum_{j\in \mathcal{C}_r}e^{-b_{j}}p_j$.
For the loss function, RTPB can be regard as a dynamic re-weighting method based on the resistance bias $b_i$ and the relationship prediction $p_i$. And the weight for the relationship $i$ between $(s,o)$ is $\theta_i$ in Eq. \eqref{eq:biasloss}.
Because $\frac{d\theta_{i} }{d b_{i}} = 1 - \frac{e^{-b_{i}} Pr(i|s,o,I)}{\sum_{j\in \mathcal{C}_r}e^{-b_{j}} Pr(j|s,o,I)} > 0 $, the RTPB can up-weight the relationships that correspond to larger resistance bias while down-weighting the rest. In this manner, RTPB reduces the loss contribution from head relationships and highlight the tail relationships to keep a balance between head and tail. 
Take binary classification as an example, and we can visualize the loss value for the pair with a specific resistance bias value; the figure is shown in the Appendix.

Moreover, the loss value tends to be larger when the predicted distribution of $p_i$ in Eq. \eqref{eq:p_i} is close to the distribution of $e^{-b_i}= \frac{w_i^{a}}{\sum_{j \in \mathcal{C}_r} w_j^a}+ \epsilon$, which is correlated to the distribution of the predefined weight $\omega_i$ of each relationship in Eq. \eqref{eq:bias}. Because the distribution of $\omega_i$ correlates to the long-tail distribution of the training set, the resistance bias provides continuous supervision against the long-tailed distribution to avoid biased prediction.

\subsubsection{Objective with RTPB}
From another point of view, we can simply regard the $\hat{\mathbf{z}}_{s,o} = \mathbf{z}_{s,o} - \mathbf{b}$  as a joint to be optimized in the training phase and the result should be close to the baseline without RTPB. We then get the final prediction $ \mathbf{z}_{s,o} =\hat{\mathbf{z}}_{s,o} + \mathbf{b}$. In this manner, compared with the model without RTPB, we tend to judge the ambiguous predictions as belonging to another one with a larger resistance bias. Thus the classification boundary of two classes is tilted to the one with larger resistance. The tilt depends on the relative value relationship of resistance bias between the two relationship categories. For example, if we judge the label of a relationship as $i$, the $\hat{z}_{i}$ should outperform all of the rest $\{\hat{z}_j, j\neq i\}$ by a necessary classification margin $b_j - b_i$ .

\section{Experiments}
In this section, we evaluate our method on the most popular scene graph generation dataset, Visual Genome (VG). To demonstrate the effectiveness of our method, we compare it with several recent methods, including MOTIFS, VCTree, and BGNN. We also perform comprehensive ablation studies on different components and provide a discussion on different cost-sensitive methods for imbalance learning.

\begin{table*}[!ht]
	\centering
    \small
     \setlength{\tabcolsep}{1mm}
	\begin{tabular}{c|ccc|ccc|ccc}
\hline
                                        & \multicolumn{3}{c|}{Predcls} & \multicolumn{3}{c|}{SGcls} & \multicolumn{3}{c}{SGdet} \\
Models                                  & mR@20   & mR@50   & mR@100   & mR@20   & mR@50  & mR@100  & mR@20   & mR@50  & mR@100  \\ \hline
KERN\ddag\cite{chen2019knowledge}       & -       & 17.7    & 19.2     & -       & 9.4    & 10.0    & -       & 6.4    & 7.3     \\
PCPL\ddag\cite{yan2020pcpl}             & -       & 35.2    & 37.8     & -       & 18.6   & 19.6    & -       & 9.5    & 11.7    \\
GPS-Net\ddag\cite{lin2020gps}           & 17.4    & 21.3    & 22.8     & 10.0    & 11.8   & 12.6    & 6.9     & 8.7    & 9.8     \\
BGNN\cite{li2021bipartite}              & -       & 30.4    & 32.9     & -       & 14.3   & 16.5    & -       & 10.7   & 12.6    \\ \hline
MOTIFS\dag\cite{Zellers_2018_CVPR}      & 13.6    & 17.2    & 18.6     & 7.8     & 9.7    & 10.3    & 5.6     & 7.7    & 9.2     \\
MOTIFS+TDE\cite{Tang2020_unbiased}      & 18.5    & 24.9    & 28.3     & 11.1    & 13.9   & 15.2    & 6.6     & 8.5    & 9.9     \\
MOTIFS+DLFE\cite{chiou2021recovering}   & 22.1    & 26.9    & 28.8     & 12.8    & 15.2   & 15.9    & 8.6     & 11.7   & 13.8    \\
MOTIFS + \textbf{RTPB(CB)}              & 28.8    & 35.3    & 37.7     & 16.3    & 19.4   & 20.6    & 9.7     & 13.1   & 15.5    \\ \hline
VCTree\dag\cite{VCTree_tang}            & 13.4    & 16.8    & 18.1     & 8.5     & 10.8   & 11.5    & 5.4     & 7.4    & 8.6     \\
VCTree+TDE\cite{Tang2020_unbiased}      & 17.2    & 23.3    & 26.6     & 8.9     & 11.8   & 13.4    & 6.3     & 8.6    & 10.3    \\
VCTree + DLFE\cite{chiou2021recovering} & 20.8    & 25.3    & 27.1     & 15.8    & 18.9   & 20.0    & 8.6     & 11.8   & 13.8    \\
VCTree+\textbf{RTPB(CB)  }              & 27.3    & 33.4    & 35.6     & \textbf{20.6}    & \textbf{24.5}   & \textbf{25.8}    & 9.6     & 12.8   & 15.1    \\ \hline
DTrans                                  & 15.1    & 19.3    & 21.0     & 9.9     & 12.1   & 13.0    & 6.6     & 9.0    & 10.8    \\
DTrans+\textbf{RTPB(CB) }               & \textbf{30.3}    & \textbf{36.2}    & \textbf{38.1 }    & 19.1    & 21.8   & 22.8    & \textbf{12.7 }   & \textbf{16.5}   & \textbf{19.0}    \\
DTrans+\textbf{RTPB(VB)}                & 25.5    & 31.1    & 33.3     & 17.3    & 20.1   & 21.3    & 11.9    & 15.7   & 18.4    \\
DTrans+\textbf{RTPB(PB)}                & 17.4    & 21.6    & 23.1     & 11.9    & 14.0   & 14.7    & 7.7     & 10.1   & 11.9    \\
DTrans+\textbf{RTPB(EB)}                & 22.7    & 26.7    & 28.4     & 15.2    & 17.4   & 18.2    & 11.0    & 14.1   & 16.1    \\ \hline
\end{tabular}
	\caption{ The performance on VG~\cite{krishna2016visual} under graph constraints setting. \dag~indicates the results reproduced using the code of \cite{Tang2020_unbiased}. \ddag~models are with VGG16 backbone \cite{simonyan2014very}, while others are with ResNeXt-101-FPN backbone \cite{Lin_2017_ICCV}. (CB), (VB), (PB), and (EB) indicate the count resistance bias, the valid resistance bias, the pair resistance bias, and the estimated resistance bias, respectively. 
	}
	\label{tab:sgg-mr}
\end{table*}

\subsection{Dataset and Evaluation Metrics}

\subsubsection{Dataset} 
We perform extensive experiments on Visual Genome (VG)~\cite{krishna2016visual} dataset. Similar to previous work~\cite{Xu2017,Zellers_2018_CVPR,Tang2020_unbiased}, we use the widely adapted subset, VG150, which consists of the most frequent 150 object categories and 50 predicate categories. The original split only has training set (70\%) and test set (30\%). We follow \cite{Tang2020_unbiased} to sample a 5k validation set for parameter tuning.

\subsubsection{Evaluation Metrics}
For a fair comparison, we follow previous work~\cite{Zellers_2018_CVPR,Tang2020_unbiased,li2021bipartite} and evaluate the proposed method on three sub-tasks of scene graph generation as follows.
\begin{enumerate}
	\item Predicate Classification (\textbf{PredCls}). On this task, the input includes not only the image, but the bounding boxes in the image and their labels;
	\item Scene Graph Classification (\textbf{SGCls}). It only gives the bounding boxes;
	\item Scene Graph Detection (\textbf{SGDet}). On this task, no additional information other than the raw image will be given.
\end{enumerate}
For each sub-task, previous works use Recall@k (or R@k) to report the performance of scene graph generation~\cite{Lu16,Xu2017}, which indicates the recall rate of ground-truth relationship triplets (subject-predicate-object) among the top k predictions of each test image.
Considering that the above-mentioned metric (R@k) caters to the bias caused by the long-tailed distribution of object relationships, the model can only achieve a high recall rate by predicting the frequent relationships. Therefore, it can not demonstrate the effectiveness of different models on less frequent relationships. To this end, the mean recall rate of each relationships, (mR@k), has been widely used in recent works ~\cite{Zellers_2018_CVPR,VCTree_tang,Tang2020_unbiased,li2021bipartite}. Therefore, we report the performance using the mean Recall, and the results of Recall are available in the appendix.

\subsection{Implementation Details}
We implement the proposed method using PyTorch~\cite{NEURIPS2019_9015}. For a fair comparison, we use the object detection model similar to \cite{Tang2020_unbiased}, i.e., a pretrained Faster RCNN with ResNeXt-101-FPN as the backbone network. The object detector is fine-tuned on the VG dataset. Then, we froze the object detector and train the rest parts of the SGG model.
 For the DTrans, the number of object encoder layers is $n_o = 4$ and the number of relationship encoder layers is $n_r = 2$. For the proposed resistance bias, we use $a = 1$ and $\epsilon =0.001$ if not otherwise stated. For the background relationship, the bias is a constant value $log\frac{1}{\|\mathcal{C}_r\|} = log\frac{1}{50}$, where the $\mathcal{C}_r$ is the set of relationship labels. We perform our experiments using a single NVIDIA V100 GPU. We train the DTrans model for 18000 iterations with batch size 16. Specifically, it takes around 24 hours for the SGDet task, and less than 12 hours for the PredCls/SGCls task. For the SGDet task, we use all pairs of objects for training instead of those pair of objects with overlap. For other experimental settings, we always keep the same with the previous work~\cite{Tang2020_unbiased}.

\subsection{Comparison with Recent Methods}
\begin{figure}[!ht]
	\centering
	\includegraphics[width=\linewidth]{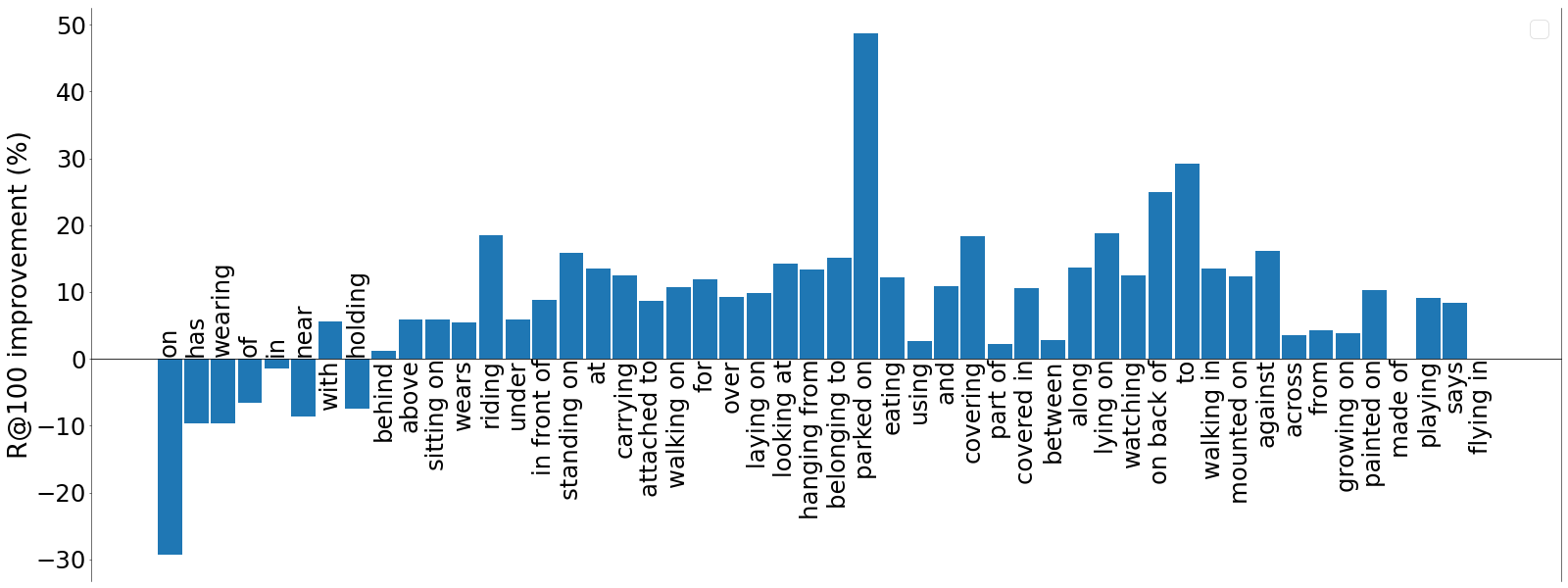} 
	\caption{The improvements of Recall@100 on the SGDet task using RTPB(CB). The number of relationships decreases from left to right. It shows that RTPB achieves higher recall rate on most relationships with a sacrifice on only a small number of frequent relationships.}
	\label{fig:compR}
\end{figure}

To demonstrate the effectiveness of the proposed method, we apply RTPB on several representative methods for scene graph generation, including MOTIFS, VCTree, and our DTrans baseline model. 
As shown in Table~\ref{tab:sgg-mr}, we find that RTPB can achieve consistent improvements in the mean recall metric for all of MOTIFS, VCTree, and our DTrans. Models equipped with our RTPB achieve new state-of-the-art and out-preform previous methods by a clear margin. 
Take SGDet as an example, as shown in Figure~\ref{fig:compR}, the recall rate on most relationships has been significantly improved (DTrans w/o or w/ RTPB). Specifically, RTPB may also lead to slight performance degradation on the frequent relationships without over-fitting on the head categories. Therefore, we see a clear improvement when using the mean recall rate as the evaluation metric, which demonstrates RTPB can build a better-balanced model for unbiased SGG.

Among four instance of resistance bias, CB performs the best on mR because it simply calculates the relationship distribution based on the whole dataset. VB/PB/EB are proposed based on more critical conditions. Although we believe that adding more critical conditions would lead to more practical prior bias, currently, limited data are not sufficient to obtain proper empirical bias under the given conditions.

\begin{figure}[!ht]
	\begin{minipage}[t]{0.49\linewidth}
		\centering
		\includegraphics[width=\linewidth]{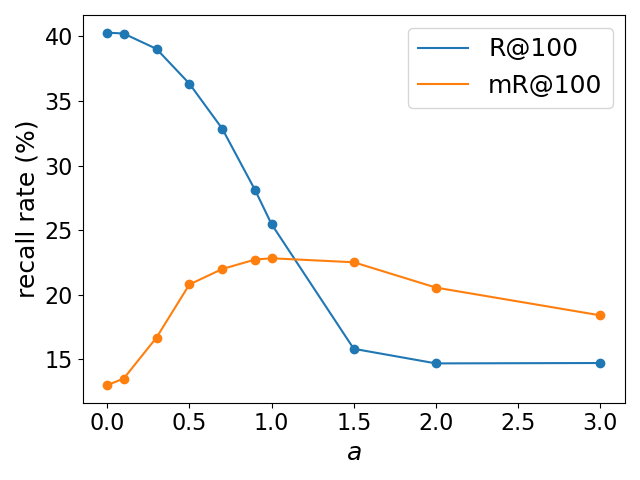}
		{(a)}
	\end{minipage}
	\begin{minipage}[t]{0.49\linewidth}
		\centering
		\includegraphics[width=\linewidth]{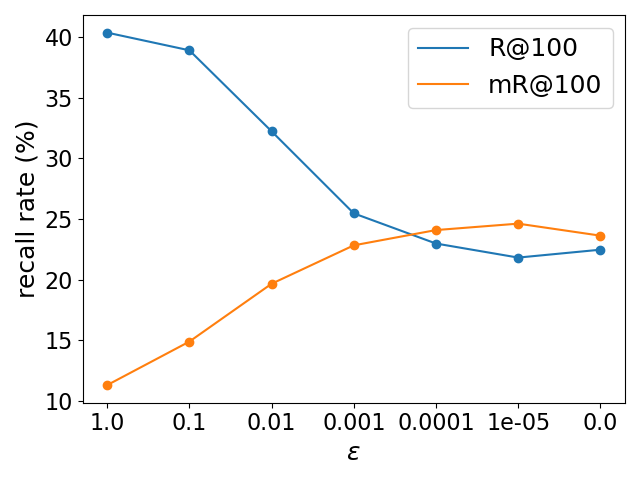}
		{(b)}
	\end{minipage}
	\caption{(a) the influence of different $a$. (b) the influence of different $\epsilon$.}
	\label{fig:param}
\end{figure}

\subsection{Ablation Studies}
In this subsection, we evaluate the influence of important hyper-parameters ($a$ and $\epsilon$) and an extended version of resistance bias, i.e., soft resistance bias. 
Besides, we compare our RTPB with several conventional imbalance learning methods on the SGG task.

\subsubsection{Hyper-parameters for Resistance Bias}

Two important hyper-parameters in RTPB, $a$ and $\epsilon$ are used to control the distribution of the resistance bias. Specifically, $a >0$ is used to adjust the divergence of the bias, while a small $\epsilon \in [0,1]$ is used to control the maximum relative difference. 
We perform experiments on the SGCls task to evaluate the influence of  $a$ and $\epsilon$ in the RTPB. As shown in Figure~\ref{fig:param}(a), when increasing $a$, the overall recall rate decreases and the mean recall rate increases; when $a > 1.0$, the mean recall rate also decreases. Therefore, we use $a=1.0$ in our experiments. As shown in Figure~\ref{fig:param}(b), we find that $\epsilon = 1e-3$ achieves a good trade-off between the overall recall rate and the mean recall rate. Furthermore, we find out that when either $a = 0$ or $\epsilon \le 1$, the performance is very close to the baseline method because the resistance biases for different relationships are close to each other. 


\begin{figure}[!ht]
	\centering
	\includegraphics[width=\linewidth]{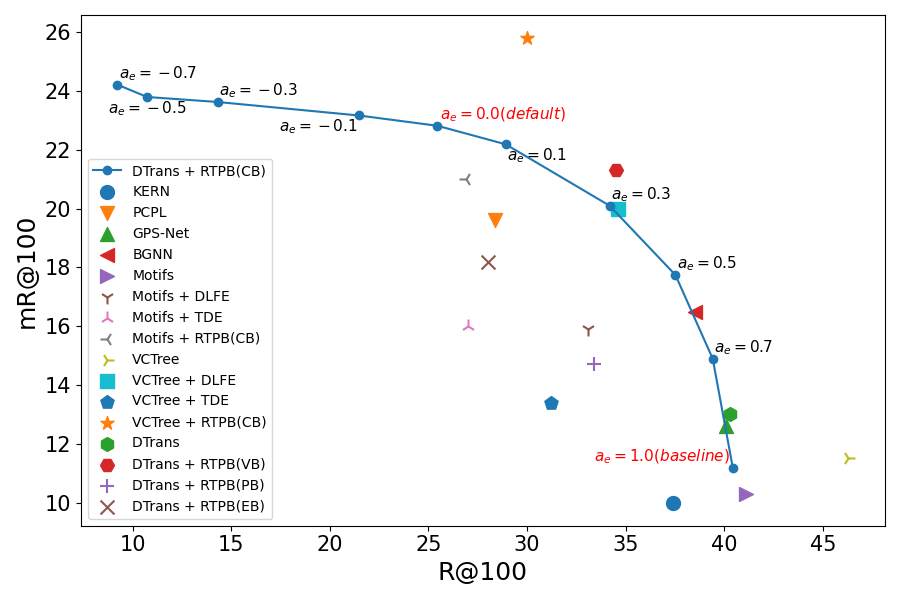}
	\caption{The result of inference with different $a_e$.}
	\label{fig:esb}
\end{figure}

\subsubsection{Soft Resistance Bias}
To better understand how RTPB addresses the trade-off between head and tail relationships, we introduce a soft version of resistance bias during inference phase as follows.
\begin{equation}
	b_i^e = -log (\frac{w_i^{a_e}}{\sum_{j \in \mathcal{C}_r} w_j^{a_e}}+ \epsilon),
\end{equation}
where $a_e \leq a$, and $b_i^e$ is the smooth version of resistance bias. By applying a smoothed resistance bias in the inference phase, we can evaluate different impacts of RTPB on the trained model. Following the analysis of the resistance bias for adjusting the classification boundary, a smooth resistance bias reduces the margin areas of head categories from $a log\frac{w_i}{w_j}$ to $(a -a_e)log\frac{w_i}{w_j}$. Specifically, 1) when $a_e= 0$, the result is same as the original one; and 2) when $a_e = a =1$, the result is close to the baseline without RTPB, i.e., Motifs~\cite{Zellers_2018_CVPR} and DTrans.
The experimental results of different $a_e$  are shown in Figure~\ref{fig:esb}. The model used for evaluation is trained with CB and $a = 1$.
As the $a_e$ decreases, the classification margin of head categories keeps becoming larger. Therefore, the model performs worse on the few head categories but performs better on the rest tail categories. For the evaluation result, this means a consistent improvement in mean recall metric as shown in Figure~\ref{fig:esb}. Besides, the model trained with RTPB can achieve similar performance as many current SGG methods with different $a_e$. Thus RTPB can be regarded as a general baseline for  different unbiased SGG methods.

\subsection{Other Cost-sensitive Methods}
As shown in Table~\ref{tab:debias}, we compare our method with the following cost-sensitive methods on the SGCls task, using our DTrans as the backbone. 
1) \textbf{Re-weighting Loss}: using the fraction of the count of each class as the weight of loss. 
2) \textbf{Class Balanced Re-weighting Loss} \cite{cui2019class}: using the effective number of samples to re-balance the loss. 
3) \textbf{Focal Loss} \cite{Lin_2017_ICCV}: using the focal weight to adjust the losses for well-learned samples and hard samples. 
4) \textbf{Label Distribution-Aware Margin Loss} \cite{cao2019learning}: using the prior margin for each class to balance the model. 

We evaluate these methods on the SGCls task with our DTrans. As shown in Table~\ref{tab:debias}, our method outperforms the rest of the methods by a clear margin in the mean recall.
\begin{table}
\centering
	\begin{tabular}{c|ccc}
\hline
 Loss & mR@20         & mR@50         & mR@100        \\ \hline
Baseline                                            & 9.9           & 12.2          & 13.0          \\
$\mathcal{L}_{Reweight}$                            & 9.9           & 14.3          & 16.8          \\
$\mathcal{L}_{ClsBal}$                              & 12.8          & 15.5          & 17.0          \\
$\mathcal{L}_{focal}$                               & 9.1           & 11.3          & 12.1          \\
$\mathcal{L}_{LDAM}$                                & 8.4           & 10.3          & 10.8          \\
RTPB(CB)                                            & \textbf{19.1} & \textbf{21.8} & \textbf{22.8} \\ \hline
\end{tabular}
	\caption{Comparison with other cost-sensitive methods.}
	\label{tab:debias}
\end{table}



\section{Conclusion}
In this paper, we propose a novel resistance training strategy using prior bias (RTPB) for unbiased SGG.
Experimental results on the popular VG dataset demonstrate that our RTPB achieves a better head-tail trade-off for the SGG task than existing counterparts. We also devise a novel transformer-based contextual encoding structure to encode global information for visual objects and relationships. We obtain significant improvements over recent state-of-the-art approaches, and thus set a new baseline for unbiased SGG.
\section*{Supplementary Material}
\subsection*{Visualization of the loss}
The softmax CE with our resistance bias is calculated as follows:
\begin{align}
	\mathcal{\hat{L}}_i 
	& = -log(\hat{p}_i)
	= -log \frac{e^{z_{i} - b_{i}}}{\sum_{j\in \mathcal{C}_r}e^{z_{j} - b_{j}}}
	\nonumber\\
	& = -log \frac{e^{z_i}}{\sum_{k\in \mathcal{C}_r}e^{z_k}} \frac{e^{-b_{i}}{\sum_{k\in \mathcal{C}_r}e^{z_k}}}{\sum_{j\in \mathcal{C}_r}e^{z_j-b_{j}}}\nonumber\\
	& = -log \frac{e^{z_i}}{\sum_{k\in \mathcal{C}_r}e^{z_k}} + b_i + log \sum_{j\in\mathcal{C}_r}\frac{e^{-b_j}e^{z_j}}{\sum_{k\in \mathcal{C}_r}e^{z_k}} 
	\nonumber\\
	& = -log~p_i + b_i + log \sum_{j\in\mathcal{C}_r}{e^{-b_j}p_j} \nonumber\\
	& = \mathcal{L}_i + \theta_{i}, \label{eq:biasloss:supp}\nonumber\\
\end{align}
where $\mathcal{L}_i$ is the vanilla CE, and $~ \theta_{i}  =  b_{i}  + log \sum_{j\in \mathcal{C}_r}e^{-b_{j}}p_j$ is the extra weight of loss value, which is correlated to the resistance bias value for each type. Specifically, the extra weight $\theta_i$ increase the loss value when the resistance bias is large.

\begin{figure}[ht!]
	\centering
	\includegraphics[width=0.9\columnwidth]{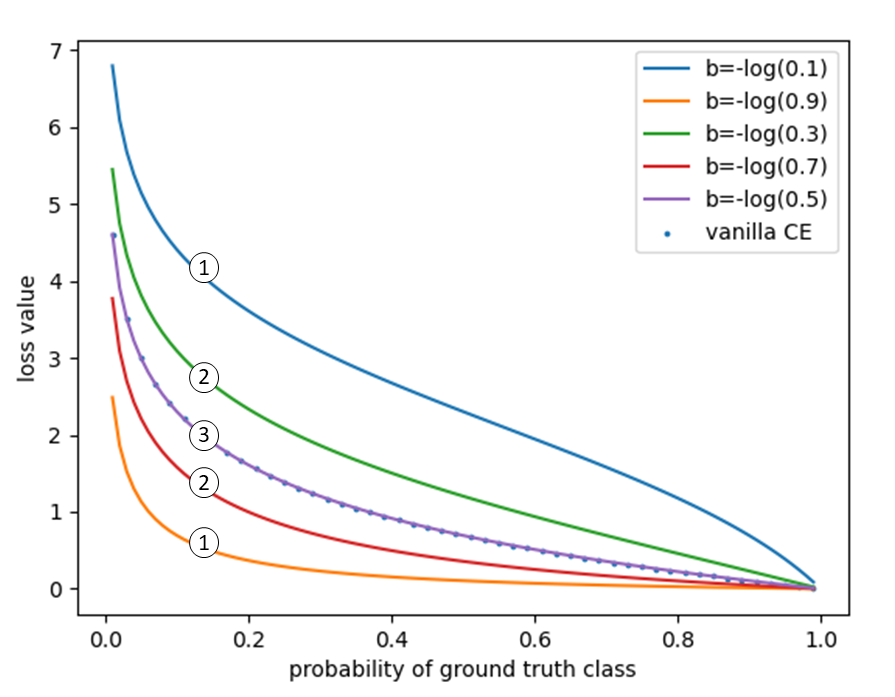} 
	\caption{The loss value for cross-entropy with our RTPB. This figure shows three pairs of loss with different resistance bias. In the pair \ding{172}, the resistance bias is set to be $-log(0.9)$ and $-log(0.1)$. $-log(0.7)$ and $-log(0.1)$ is the bias value for the second pair, \ding{173}. In the last pair, \ding{174}, two classes take the same resistance bias value, and the loss value is equivalent to the vanilla CE \cite{Mannor2005Cross}. }
	\label{fig:loss}
\end{figure}

Take binary classification as an example, we can visualize the loss value for the two classes with specific resistance bias values. The loss values for several pair of resistance bias values are shown in Figure \ref{fig:loss}. 

\subsection*{Experiment Results}
\begin{figure}[!ht]
	\centering
	\includegraphics[width=0.95\linewidth]{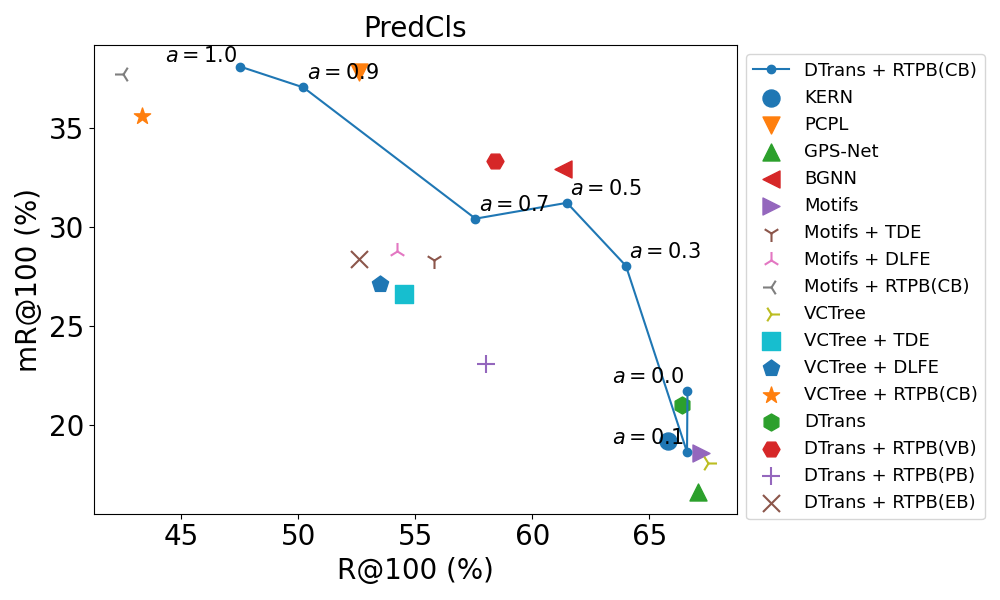} 
	\caption{The distribution of R@100 and mR@100 in PredCls task.}
	\label{fig:R_mR_predcls}
\end{figure}
\begin{figure}[!ht]
	\centering
	\includegraphics[width=0.95\linewidth]{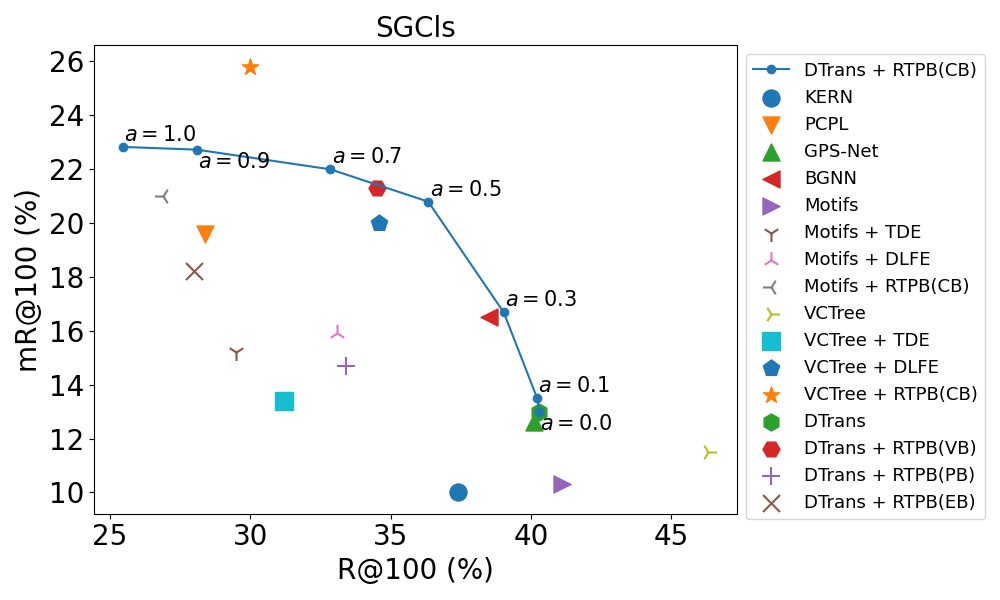} 
	\caption{The distribution of R@100 and mR@100 in SGCls task.}
	\label{fig:R_mR_sgcls}
\end{figure}
\begin{figure}[!ht]
	\centering
	\includegraphics[width=0.95\linewidth]{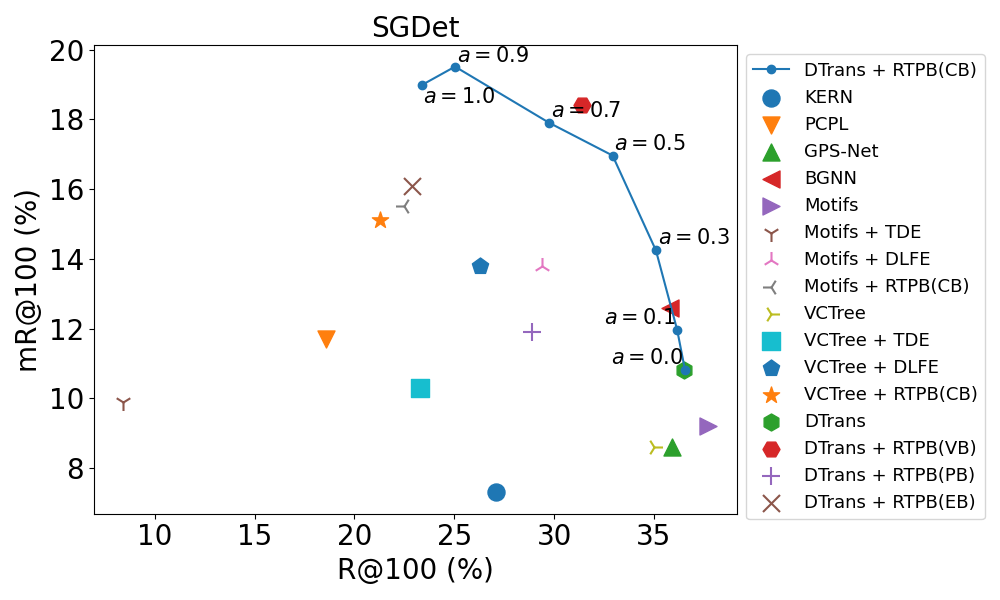} 
	\caption{The distribution of R@100 and mR@100 in SGDet task.}
	\label{fig:R_mR_sgdet}
\end{figure}
Table \ref{tab:sgg-r-mr} shows the Recall results and mean Recall results on the VG dataset \cite{krishna2016visual}.
\begin{table*}[!ht]
	\centering
	\small
	\begin{tabular}{c|cc|cc|cc}
		\hline
		& \multicolumn{2}{c|}{Predcls} & \multicolumn{2}{c|}{SGcls} & \multicolumn{2}{c}{SGdet} \\
		Models                         & mR@50/100     & R@50/100     & mR@50/100    & R@50/100    & mR@50/100    & R@50/100    \\ \hline
		KERN\ddag\cite{chen2019knowledge}                      & 17.7/19.2     & 67.6/65.8    & 9.4/10.0     & 36.7/37.4   & 6.4/7.3      & 29.8/27.1   \\
		PCPL\ddag\cite{yan2020pcpl}                      & 35.2/37.8     & 50.8/52.6    & 18.6/19.6    & 27.6/28.4   & 9.5/11.7     & 14.6/18.6   \\
		GPS-Net\ddag\cite{lin2020gps}                        & 21.3/22.8   & 66.9/68.8    & 11.8/12.6      & 39.2/40.1   &8.7/9.8      & 28.4/31.7   \\
		BGNN\cite{li2021bipartite}                           & 30.4/32.9     & 59.2/61.3    & 14.3/16.5    & 37.4/38.5   & 10.7/12.6    & 31.0/35.8   \\ \hline
		MOTIFS\dag\cite{Zellers_2018_CVPR}                    & 17.2/18.6     & 65.4/67.2    & 9.7/10.3     & 40.3/41.1   & 7.7/9.2      & 33.0/37.7   \\
		MOTIFS+TDE\cite{Tang2020_unbiased}                        & 24.9/28.3     & 50.8/55.8    & 13.9/15.2    & 27.2/29.5   & 8.5/9.9    & 7.4/8.4  \\
		MOTIFS+DLFE\cite{chiou2021recovering}                    & 26.9/28.8     & 52.5/54.2    & 15.2/15.9    & 32.3/33.1   & 11.7/13.8    & 25.4/29.4   \\
		MOTIFS + \textbf{RTPB(CB)}               & 35.3/37.7     & 40.4/42.5    & 20.0/21.0    & 26.0/26.9   & 13.1/15.5    & 19.0/22.5   \\ \hline
		VCTree\dag\cite{VCTree_tang}                    & 16.8/18.1     & 65.8/67.5    & 10.8/11.5    & 45.4/46.3   & 7.4/8.6      & 31.0/35.0   \\
		VCTree+TDE\cite{Tang2020_unbiased}                        & 23.3/26.6     & 49.9/54.5    & 11.8/13.4 & 28.8/31.2   & 8.6/10.3    & 19.6/23.3  \\
		VCTree + DLFE\cite{chiou2021recovering}                  & 25.3/27.1     & 51.8/53.5    & 18.9/20.0    & 33.5/34.6   & 11.8/13.8    & 22.7/26.3   \\
		VCTree+\textbf{RTPB(CB)  }               & 33.4/35.6     & 41.2/43.3    &\textbf{ 24.5/25.8}    & 28.7/30.0   & 12.8/15.1    & 18.1/21.3   \\ \hline
		DTrans                         & 19.3/21.0     & 64.4/66.4    & 12.1/13.0    & 39.4/40.3   & 9.0/10.8     & 32.1/36.5   \\
		DTrans+\textbf{RTPB(CB) }                & \textbf{36.2/38.1}    & 45.6/47.5    & 21.8/22.8    & 24.5/25.5   & \textbf{16.5/19.0}    & 19.7/23.4   \\
		DTrans+\textbf{RTPB(VB)}   & 31.1/33.3                    & 56.5/58.4                  & 20.1/21.3                 & 33.6/34.5 & 15.7/18.4 & 27.3/31.4 \\
		DTrans+\textbf{RTPB(PB)}         & 21.6/23.1        & 56.2/58.0       & 14.0/14.7    & 32.6/33.4   & 10.1/11.9    & 24.6/28.9   \\
		DTrans+\textbf{RTPB(EB)}                 & 26.7/28.4     & 50.6/52.6    & 17.4/18.2    & 27.0/28.0   & 14.1/16.1    & 19.1/22.9   \\
		
		\hline
	\end{tabular}
	\caption{ The performance on VG~\cite{krishna2016visual} under graph constraints setting. \dag~indicates the results reproduced using the code of \cite{Tang2020_unbiased}. \ddag~models are with VGG16 backbone \cite{simonyan2014very}, while others are with ResNeXt-101-FPN backbone \cite{Lin_2017_ICCV}. (CB), (VB), (PB) and (EB) indicate the count resistance bias, the valid resistance bias, the pair resistance bias, and the estimated resistance bias, respectively.}
	\label{tab:sgg-r-mr}
\end{table*}
To clearly compare all methods, we visualize the results on the PredCls task, SGCls task, and SGDet task in Figure \ref{fig:R_mR_predcls}, \ref{fig:R_mR_sgcls} and \ref{fig:R_mR_sgdet}, respectively.  As shown in Figure \ref{fig:R_mR_predcls}, \ref{fig:R_mR_sgcls} and \ref{fig:R_mR_sgdet}, models trained with RTPB achieve higher mean recall than previous state-of-the-art methods. Besides, the model trained with different $a$ for RTPB provide examples for head-tail trade-off in SGG.

Moreover, in addition to Recall and mean Recall result with graph constraint, we also report the results without graph constraint of our methods in Table \ref{tab:sgg-r-mr-ng}. As we can observe, RTPB also significantly improves the mean Recall of corresponding methods. Besides, DTrans+RTPB(CB) performs the best on nearly all mean Recall. The Recall is dropped by using RTPB. This is because the less frequently seen relationships have been highlighted, where most ground truths are frequently seen. Therefore, it is commonly seen in unbiased SGG strategies that an increase of mean Recall would be followed with a decrease of Recall \cite{Tang2020_unbiased}. 

\begin{table*}[!ht]
	\centering
	\small
	\begin{tabular}{c|cc|cc|cc}
		\hline
		& \multicolumn{2}{c|}{Predcls}   & \multicolumn{2}{c|}{SGcls}     & \multicolumn{2}{c}{SGdet}     \\
		Models                                  & mR@50/100          & R@50/100  & mR@50/100          & R@50/100  & mR@50/100          & R@50/100  \\ \hline
		MOTIFS\dag\cite{Zellers_2018_CVPR}      & 33.8/46.0          & 81.7/88.9 & 19.9/26.8          & 50.3/54.0 & 13.2/17.9          & 37.2/44.1 \\
		MOTIFS+TDE\S~\cite{Tang2020_unbiased}      & 29.0/38.2          & -         & 16.1/21.1          & -         & 11.2/14.9          & -         \\
		MOTIFS+DLFE\S~\cite{chiou2021recovering}   & 30.0/45.8          & -         & 25.6/32.0          & -         & 18.1/23.0          & -         \\
		MOTIFS + \textbf{RTPB(CB)}              & 48.0/58.9          & 62.8/74.9 & 27.2/33.3          & 39.6/46.5 & 17.5/22.5          & 25.7/32.6 \\ \hline
		VCTree\dag\cite{VCTree_tang}            & 34.6/46.9          & 82.5/89.5 & 22.2/30.7          & 56.3/60.6 & 12.6/16.8          & 33.6/39.9 \\
		VCTree+TDE\S~\cite{Tang2020_unbiased}      & 32.4/41.5          & -         & 19.1/25.5          & -         & 11.5/15.2          & -         \\
		VCTree + DLFE\S~\cite{chiou2021recovering} & 44.6/56.8          & -         & 31.4/38.8          & -         & 17.5/22.5          & -         \\
		VCTree+\textbf{RTPB(CB)  }              & 46.4/57.8          & 62.6/74.9 & \textbf{34.0/41.1} & 45.1/53.7 & 17.3/21.9          & 24.5/31.4 \\ \hline
		DTrans                                  & 36.6/48.8          & 80.5/87.9 & 23.2/29.7          & 49.5/53.3 & 14.4/19.6          & 36.3/43.0 \\
		DTrans+\textbf{RTPB(CB) }               & \textbf{52.6/63.5} & 70.0/80.7 & 31.6/37.1          & 42.0/48.5 & \textbf{22.4/27.5} & 27.5/35.3 \\
		DTrans+\textbf{RTPB(VB)}                & 47.5/59.7          & 76.5/85.3 & 30.6/36.9          & 46.6/51.4 & 21.3/27.2          & 32.5/39.8 \\
		DTrans+\textbf{RTPB(PB)}                & 38.6/50.4          & 79.2/87.7 & 25.4/31.9          & 48.4/52.8 & 16.6/21.7          & 33.4/41.1 \\
		DTrans+\textbf{RTPB(EB)}                & 42.5/52.6          & 76.4/85.9 & 27.5/33.0          & 46.1/51.6 & 19.9/24.7          & 30.4/38.2 \\ \hline
	\end{tabular}
	\caption{ The performance on VG~\cite{krishna2016visual} without graph constraint. \dag, \ddag, (CB), (VB), (PB) and (EB) are with the same meaning as in Table \ref{tab:sgg-r-mr}. $\S~$ denote the results are reported by \cite{chiou2021recovering} 
	}
	\label{tab:sgg-r-mr-ng}
\end{table*}
\subsection*{Cost-sensitive Methods}

\begin{table*}[!ht]
	\centering
	\begin{tabular}{c|cc|cc|cc}
		\hline
		& \multicolumn{2}{c|}{Predcls} & \multicolumn{2}{c|}{SGcls} & \multicolumn{2}{c}{SGdet} \\
		Models                         & mR@50/100     & R@50/100     & mR@50/100    & R@50/100    & mR@50/100    & R@50/100    \\ \hline
		Baseline   & 19.3/21.0     & 64.4/66.4    & 12.1/13.0    & 39.4/40.3   & 9.0/10.8     & 32.1/36.5           \\
		$\mathcal{L}_{Reweight}$  & 35.1/\textbf{38.8} & 26.8/30.6 & 14.3/16.8  & 13.6/15.6 & 2.0/2.6 & 4.0/5.1           \\
		$\mathcal{L}_{ClsBal}$     & 19.9/21.9 & 64.5/66.3 & 15.5/17.0  & 38.8/39.7 & 10.7/12.8 & 31.8/36.2           \\
		$\mathcal{L}_{focal}$     & 19.9/21.7 & 63.2/65.6 & 11.3/12.1  & 38.6/39.8 & 8.3/10.0 & 29.6/33.9          \\
		$\mathcal{L}_{LDAM}$      & 15.3/16.6 & 65.1/66.9 & 10.3/10.8    & 39.6/40.4  & 7.0/8.6 & 32.3/37.1        \\
		RTPB(CB)      & \textbf{36.2}/38.1    & 45.6/47.5    & \textbf{21.8/22.8}    & 24.5/25.5   & \textbf{16.5/19.0}    & 19.7/23.4 \\ \hline
	\end{tabular}
	\caption{Comparison of Recall rate and mean Recall rate (with graph constraint) with other cost-sensitive methods.}
	\label{tab:debias_supp}
\end{table*}
\begin{table*}[!ht]
	\centering
	\begin{tabular}{c|cc|cc|cc}
		\hline
		& \multicolumn{2}{c|}{Predcls} & \multicolumn{2}{c|}{SGcls} & \multicolumn{2}{c}{SGdet} \\
		Models                   & mR@50/100     & R@50/100     & mR@50/100    & R@50/100    & mR@50/100    & R@50/100    \\ \hline
		Baseline                 & 19.3/21.0     & 64.4/66.4    & 12.1/13.0    & 39.4/40.3   & 9.0/10.8     & 32.1/36.5   \\
		$\mathcal{L}_{Reweight}$ & 41.2/50.0     & 40.4/53.0    & 16.4/21.1    & 20.3/26.7   & 2.7/3.7      & 5.4/7.0     \\
		$\mathcal{L}_{ClsBal}$   & 38.4/51.5     & 80.3/87.9    & 26.9/34.7    & 49.0/52.9   & 16.9/22.3    & 35.9/42.7   \\
		$\mathcal{L}_{focal}$    & 27.0/36.9     & 76.6/84.8    & 18.4/24.2    & 46.9/51.4   & 12.4/16.8    & 33.2/39.5   \\
		$\mathcal{L}_{LDAM}$     & 33.6/44.9     & 80.2/87.9    & 22.6/28.9    & 49.0/52.9   & 13.0/17.9    & 34.8/42.1   \\
		RTPB(CB)                 & \textbf{52.6/63.5}     & 70.0/80.7    & \textbf{31.6/37.1}    & 42.0/48.5   & \textbf{22.4/27.5}    & 27.5/35.3   \\ \hline
	\end{tabular}
	\caption{Comparison of Recall rate and mean Recall rate (without graph constraint) with other cost-sensitive methods.}
	\label{tab:debias_ng}
\end{table*}
We compare our RTPB with the following cost-sensitive methods. 

1) \textbf{Re-weighting Loss}: This strategy uses the fraction of the count of each class to weigh the loss value as follows,
\begin{equation}
	\mathcal{L}_{Reweight}(i) = \frac{1}{n_i}\mathcal{L}_i, 
\end{equation}
where $n_i$ is the number of all samples of $i$-th class in training set. For SGG, $n_i$ is set to be the number of relationship $i$.

2) \textbf{Class Balanced Re-weighting Loss} \cite{cui2019class}: This method uses the inversed effective number of samples to re-balance the loss. The loss is calculated as
\begin{equation}
	\mathcal{L}_{ClsBal}(i) = \frac{1}{n_i^{eff}}\mathcal{L}_i = \frac{1-\beta}{1-\beta^{n_i}}\mathcal{L}_i,
\end{equation}
where $n_i^{eff} = \frac{1-\beta}{1-\beta^{n_i}}$ is the effective number for class $i$, $n_i$ is the number of sample of class $i$. Following the setting of  \cite{cui2019class}, we set the hyper-parameter $\beta= 0.999$. 

3) \textbf{Focal Loss} \cite{Lin_2017_ICCV}: This method uses the focal weight to adjust the loss value for well-learned samples and focuses on the hard ones.
\begin{equation}
	\mathcal{L}_{focal}(p_i) = -\alpha(1-p_i)^\gamma log(p_i),
\end{equation}
where $p_i$ is the classification probability for class i. We followed the hyper-parameters ($\gamma = 2.0, \alpha= 0.25$). 

4) \textbf{Label Distribution-Aware Margin Loss} \cite{cao2019learning}: This method uses the margin for each class to balance the model. The loss value for class $i$ is
\begin{equation}
	\begin{split}
		\mathcal{L}_{LDAM}(i) = -log \frac{e^{z_i - \Delta_i}}{e^{z_i - \Delta_i} + \sum_{j\neq i}{e^{z_j}}} \\
		where ~ \Delta_i = \frac{C}{n_i^{1/4}}
	\end{split}
\end{equation}
where $C$ is the constant for the margin size and we set it as 0.5 same as \cite{cao2019learning}, and  $n_i$ is the number of sample of the corresponding class $i$ in the training set. This class-related loss can be regarded as a special form of our method if we set the bias corresponding to the ground truth as follow
\begin{equation}
	b_i =\left\{
	\begin{aligned}
		\Delta_i, ~ & i = gt   \\
		0, ~        & i\neq gt
	\end{aligned}
	\right.
	~,
\end{equation}
where $gt$ is the groundtruth of relationship.
We evaluate methods mentioned above on the VG data set\cite{krishna2016visual} with our DTrans, and find out that our RTPB outperforms the rest of the methods by a clear margin in the mean recall metric. Full results are shown in the Table \ref{tab:debias_supp}.

\subsection*{Qualitative Study}
\begin{figure*}[!ht]
	\centering
	\includegraphics[width=\linewidth]{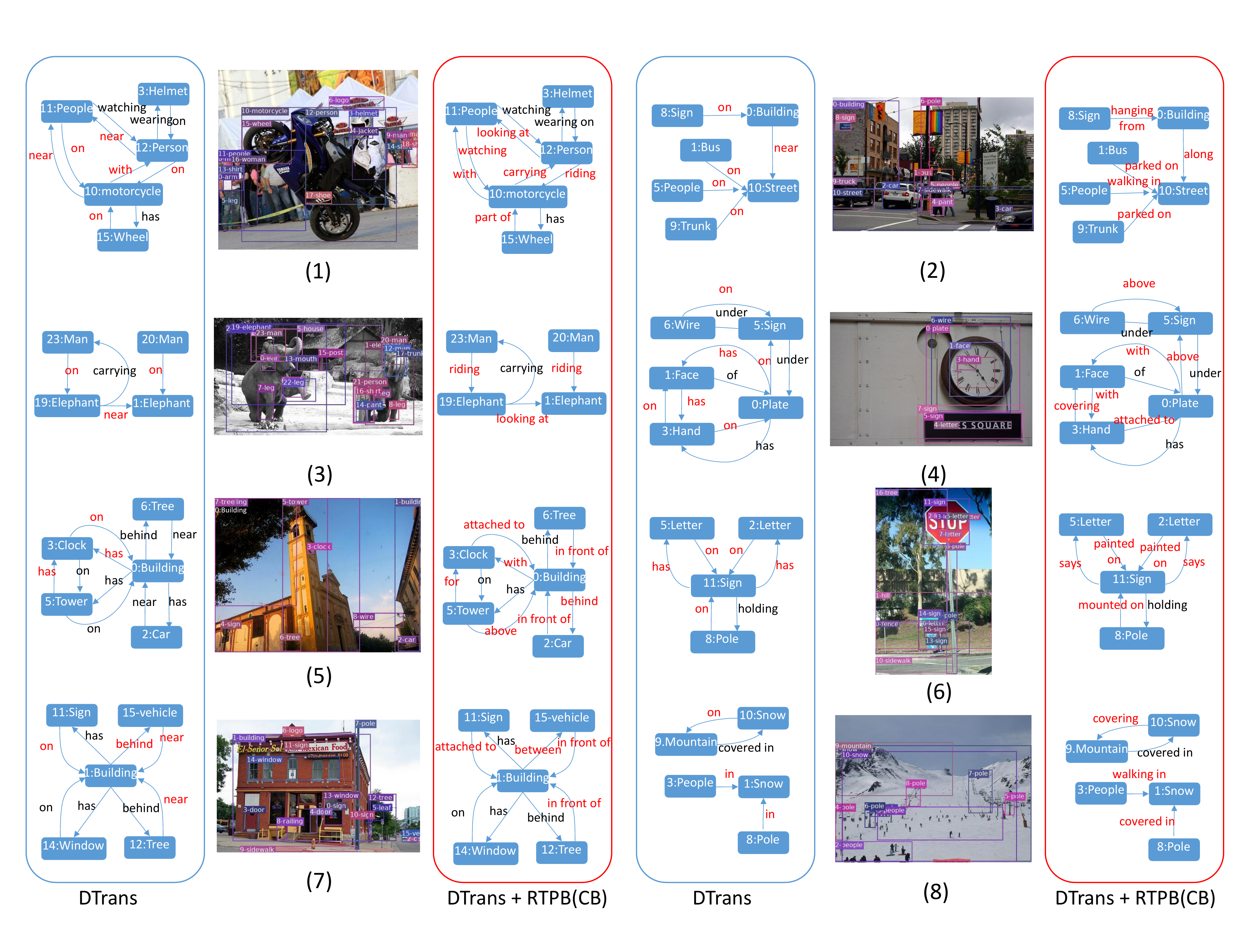} 
	\caption{This figure shows several sample of SGCls. For each sample, we present the groundtruth of object detection in the image and two scene graphs generated by different methods. The scene graphs in the blue boxes are produced by our DTrans without RTPB, and the ones in the red boxes are produced by the DTrans trained with RTPB(CB). We highlight the differences between each pair of scene graphs. Because of the space limitation, we exclude part of trivial detected objects in generated scene graphs. }
	\label{fig:sample}
\end{figure*}

We visualized several SGCls examples generated by the DTrans and the DTrans with RTPB(CB) in Figure \ref{fig:sample}. As shown in the figure, the model without RTPB prefers generalized descriptions like $on$ and $has$, which provide less valuable information. This is because these generalized relationships or the frequently seen relationships dominate the training set. However, the model trained with RTPB tends to replace these rough-grained relationships with fine-grained and more informative ones like $parked~on$, $mounted~on$, and $painted~on$, which constitutes more meaningful scene graphs. The differences between the results from the two models demonstrate the necessity of unbiased scene graph generation and the effectiveness of our proposed RTPB.
\section*{Acknowledgements}
This work was supported in part by the National Natural Science Foundation of China under Grants 61822113, 41871243, 62002090, and the Science and Technology Major Project of Hubei Province (Next-Generation AI Technologies) under Grant 2019AEA170.
Dr. Baosheng Yu is supported by ARC project FL-170100117.
Dr. Liu Liu is supported by ARC project DP-180103424.

\bibliography{ref}

\end{document}